\documentclass{article}

\usepackage{amsmath}
\usepackage{amssymb}
\usepackage{graphicx}
\usepackage{xcolor}
\usepackage{tabularx}
\usepackage{array}
\usepackage{authblk}
\usepackage[margin=1in]{geometry}
\usepackage{booktabs}
\usepackage{makecell}
\usepackage[hidelinks]{hyperref}
\usepackage{xr-hyper}
\usepackage{algorithm}
\usepackage{algpseudocode}
\usepackage{listings}

\title{Agentic self-driving microscopy benchmarks support qualification but do not necessarily generalize to unseen tasks}
\author[1]{Nathan S. Johnson}
\author[1]{Ian Abshire}
\affil[1]{Carl Zeiss Research Microscopy Solutions, 5300 Central Boulevard, Dublin, CA 94568}
\date{}

\begin{document}

\maketitle

\begin{abstract}
Large language model (LLM) agents are increasingly being developed to control a wide range of scientific characterization tools including microscopes and synchrotron beamlines. Research into agentic control of physical infrastructure is nascent and there are few well-established paradigms for how to engineer an agentic system. There are many choices to make when designing a microscopy agent, including the choice of LLM, the number of agents to use, agent responsibilities and delegation routes, retrieval-augmented generation (RAG) parameters, and more. When designing and optimizing an agentic microscope controller, researchers not only want to ensure that the agent can correctly perform known tasks but also that the agent can generalize to new tasks that it has not encountered before. In this study, we develop a benchmark and trace-logging framework that reveals a) how different choices of agent architecture impact performance at microscopy tasks and b) the limitations of benchmarks for predicting if a particular agent will perform well on unseen microscopy tasks. The framework was used to evaluate one-, two-, and three-agent graph topologies, five LLMs, RAG and context parameters, and operational constraints across 53 microscopy benchmark tests. In total, 105 agent configurations, 1,949 individual test runs, and 49,109 RAG retrievals were recorded. Direct comparisons showed clear differences in latency, token use, cost, and failure mode between configurations. RAG retrieval auditing identified knowledge fragments associated with success or failure, but pruning based on these retrospective associations did not consistently improve performance. However, surrogate models trained on agent architecture and test results did not reliably predict an agent's performance on new, unseen tasks. In a prospective comparison, a one-agent configuration with a heavily pruned RAG database produced the highest observed aggregate performance and was Pareto-optimal, although a baseline configuration with no RAG database performed approximately as well on the standardized benchmark tests. These results show that these benchmarks are useful for qualification, regression testing, diagnosis, and direct comparison, but the current heterogeneous test suite does not support a task-independent global configuration model.
\end{abstract}

\section{Introduction}
Automation is being introduced throughout the scientific process, including hypothesis generation \cite{Gottweis2026,Ghareeb2026}, sample synthesis \cite{Bran2024}, and data analysis \cite{Hollman2025}. These efforts span materials science \cite{Liang2025,Abolhasani2023}, chemistry \cite{Seifrid2022,Gary2024}, and the biological and pharmacological sciences \cite{Gottweis2026,Martin2023}. When automated decision making is combined with robotic sample handling and characterization, it forms the basis of the self-driving or autonomous laboratory \cite{Abolhasani2023,Seifrid2022,Martin2023,Gary2024}.

Microscopy remains a difficult part of this automation problem. Advanced X-ray \cite{JohnsonAugustPrototype}, electron \cite{ChenEMSeek}, synchrotron \cite{Mathur2025VISION,du2026EAA,Cherukara2026,wall2025temagentenhancingtransmission}, and scanning-probe instruments \cite{xiao2026SPMBench,Kalinin2021autoSPM,Liu2025SPM,Mandal2025} require a sequence of adaptive decisions involving alignment, parameter selection, image acquisition, and analysis. A typical workflow may require source calibration, sample centering, magnification selection, exposure-time determination, and repeated evaluation of image quality. The correct decision often depends on the sample, the instrument state, and the results of measurements made earlier in the workflow.

\begin{figure*}[t]
\centering
\includegraphics[width=\linewidth]{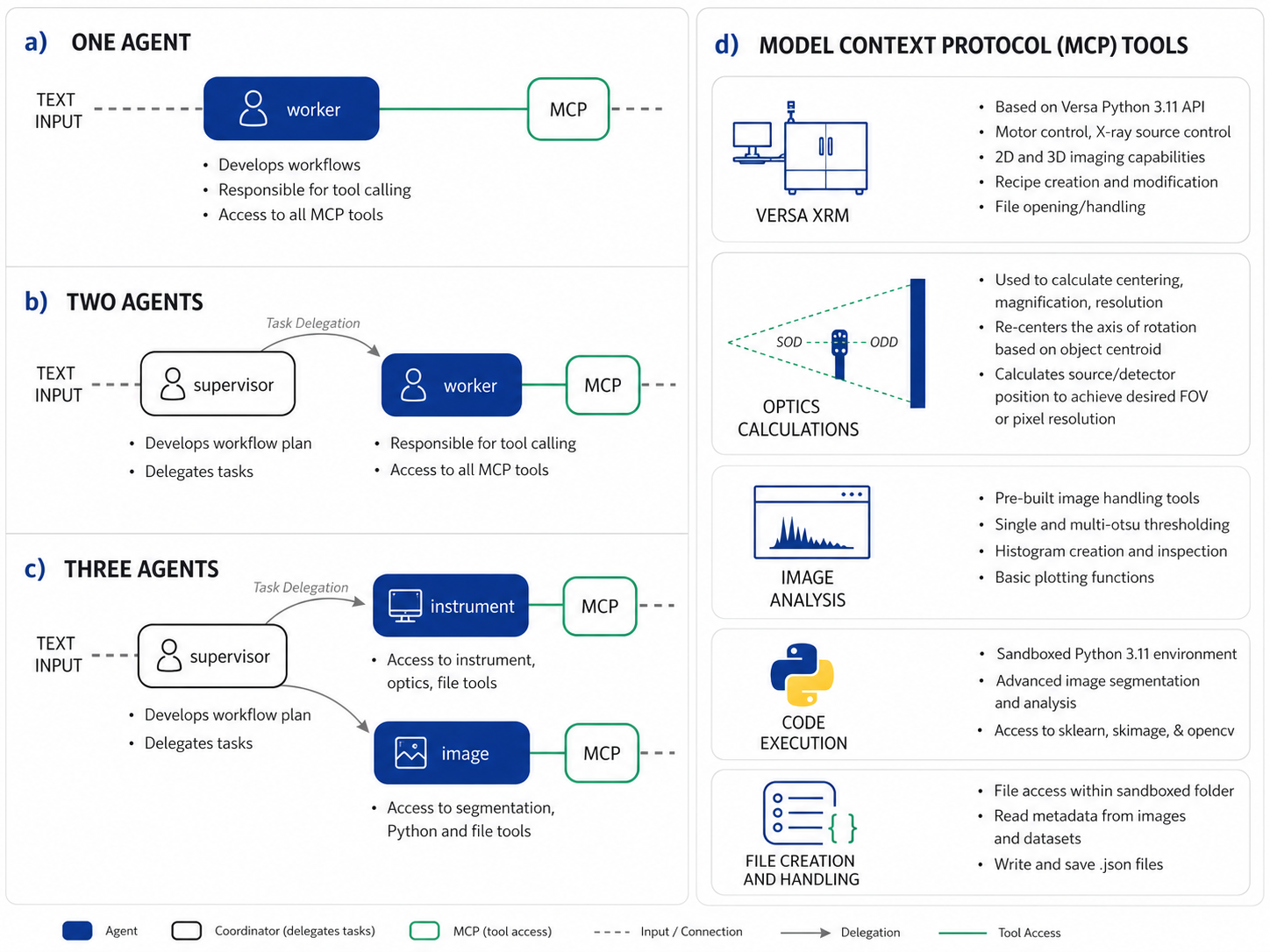}
\caption{Organization of the agent graph topologies and MCP tool surface. (a)--(c) One-, two-, and three-agent graphs and the general responsibilities assigned to each agent. (d) Tool groups exposed through six MCP servers for instrument control, recipe handling, geometry and optics calculations, image analysis, file handling, and Python execution.}
\label{fig:system_setup}
\end{figure*}

Many individual steps have already been automated. Reinforcement-learning methods have been used to tune microscope parameters and guide data collection \cite{Kalinin2021autoSPM,Liu2025SPM}. Active-learning methods have been used to prioritize measurements when time and experimental resources are limited \cite{Lookman2019,Deringer2021,Tao2021MLforMatSynth}. Microscopy data can also be segmented and analyzed using methods ranging from traditional computer vision to deep learning \cite{Helmy2023SegmentationBio,Alrfou2024SegmentationMicrostructures,Muller2024SteelSegmentation}. More recently, LLM agents have been used to coordinate longer workflows on atomic-force microscopes \cite{Mandal2025}, scanning electron microscopes \cite{ChenEMSeek}, X-ray microscopes \cite{JohnsonAugustPrototype}, transmission electron microscopes \cite{wall2025temagentenhancingtransmission}, and synchrotron beamlines \cite{corrao2025modularframeworkcollaborativehumanai}.

The design of an agentic microscope controller introduces a second optimization problem. A system designer must choose the LLM, the number and responsibilities of agents, the tool interface, the amount of conversational history retained, the use of retrieval-augmented generation (RAG), and the parameters that control retrieval and model sampling. These choices can affect success rate, latency, cost, token use, and the way that the system fails. However, it is not clear whether performance measured on one collection of microscopy tasks can be used to identify a configuration that will also perform well on a previously unseen task.

The benchmark itself also has to be designed for a specific purpose. Some tests are intended to verify that an instrument action can be completed safely. Others detect software regressions, enforce latency or tool-call limits, compare a trace against an expected workflow, or evaluate a final quantitative measurement. Quantitative correctness is the ultimate scientific endpoint, but smoke tests, regression tests, and operational limits remain necessary for qualifying an instrument-driving agent. A useful benchmark should therefore make clear what each test measures and what conclusions can be drawn from the resulting pass rate.

\begin{figure*}
\centering
\includegraphics[width=\linewidth]{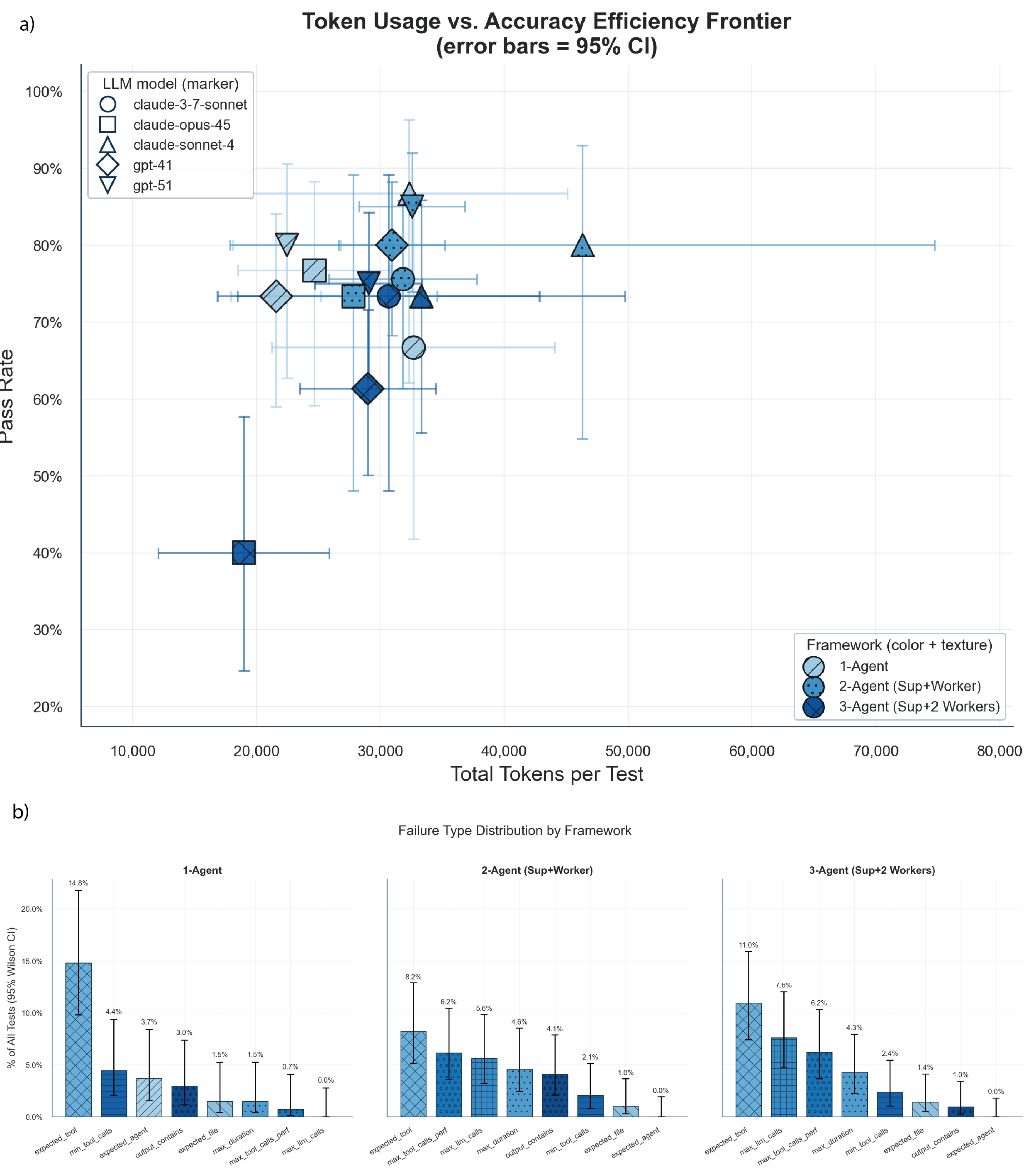}
\caption{Direct comparison of one-, two-, and three-agent topologies on the common 15-test screening set. (a) Tradeoff between composite pass rate and total tokens per test for the evaluated LLMs and graph topologies. Error bars show 95\% confidence intervals. (b) Distribution of recorded failure types for each topology.}
\label{fig:all_agent_optimizations}
\end{figure*}

In this study, we developed one-, two-, and three-agent microscope controllers connected to six Model Context Protocol (MCP) servers for instrument control, recipe handling, geometry calculations, image analysis, file handling, and Python execution. The graph topologies and tool organization are summarized in Figure~\ref{fig:system_setup}. We evaluated 105 configurations across 53 benchmark tests and linked every run to its configuration, tool trace, operational metrics, and RAG retrieval identifiers. We then used these data to ask three questions: whether agent configuration can be optimized from historical benchmark data, whether such an optimizer transfers to unseen tasks, and whether retrieval-level associations can be used to improve the RAG database. Finally, we tested selected configurations prospectively and evaluated the agent on physical microscopy workflows and quantitative measurements.

\section{Results}

\subsection{The benchmark produced a heterogeneous performance landscape}
One-, two-, and three-agent graph topologies were connected to six MCP servers that exposed tools for microscope control, recipe handling, geometry calculations, image analysis, file operations, and Python execution. Documents describing instrument procedures, sample-specific information, prior workflows, and operating guidance were embedded in RAG databases available to the agents. The system architecture is shown in Figure~\ref{fig:system_setup}, and the principal configuration variables are listed in Table~\ref{tab:hyperparameter_list}.

The benchmark contained 53 tests divided into seven suites: a six-test end-to-end workflow, 12 integration workflows, 19 atomic instrument-control tests, six geometry and recipe tests, eight regression and performance tests, one image-analysis test, and one debugging test. The tests ranged from a single instrument action to workflows that required setup, image acquisition, segmentation, geometric calculation, recipe generation, and safe shutdown. The composite pass criteria included scientific or numerical correctness together with the trace, file, and operational checks defined for each test.

Across the study, 105 agent configurations were represented by 1,949 test runs, and 49,109 RAG retrievals were recorded. Test difficulty varied substantially. Many atomic instrument tests passed for almost every functioning configuration, whereas several long integration workflows failed for most configurations. The test-level pass rates are shown in Supplementary Figure 1. The near-universal tests were useful as qualification, smoke, safety, and regression tests, but they contained little information for distinguishing among agent configurations.

\subsection{Direct architecture comparisons showed an efficiency penalty for additional agents}
The first stage compared one-, two-, and three-agent topologies on a common subset of 15 tests using the five LLMs included in the study. The results are summarized in Figure~\ref{fig:all_agent_optimizations}. Single- and two-agent systems covered a similar range of pass rates, but the one-agent systems used as many as 10,000 fewer tokens per test than corresponding two-agent systems. The three-agent topology did not provide a consistent improvement in pass rate and was not included in the subsequent broad configuration sweep.

The topology also changed the dominant failure mode. Two-agent systems more frequently exceeded limits on duration, LLM calls, or tool calls. This behavior is consistent with previous observations of agents continuing to take actions after the requested task is effectively complete \cite{Mandal2025}. Single-agent systems more often failed a trace-based check because they completed a task without calling the exact tool or following the decomposition expected by the benchmark. The difference shows why the component checks remained useful even though the headline pass rate combined all criteria.

\subsection{The surrogate interpolated known tests but did not generalize to unseen tasks}
The initial random-forest surrogate was evaluated by holding out complete agent configurations while allowing the same test identities to appear in both training and validation folds. Under this configuration-held-out evaluation, the model achieved an ROC-AUC of approximately 0.78. This result showed that the model could interpolate within the established benchmark and estimate the performance of new configurations on tasks that were already represented in the training data.

The feature-importance analysis showed why this result did not imply generalization to a new task. Test identity was the strongest predictor of the pass outcome, followed by LLM choice and the number of agents. Simple instrument-control tests passed under most configurations, while long integration workflows remained difficult. The surrogate therefore learned a substantial amount about intrinsic task difficulty rather than a task-independent configuration response.

A stricter analysis held out complete test identities and removed test identity from the design matrix. Under this evaluation, the L2-regularized logistic-regression and random-forest models each achieved an ROC-AUC of 0.55. Removing near-universally passing tests did not materially improve the task-held-out result. The configuration variables recorded in this study therefore did not provide a stable global ranking that transferred across the heterogeneous benchmark. Table~\ref{tab:surrogate_generalization} summarizes the two validation schemes.

\begin{table}
\centering
\small
\setlength{\tabcolsep}{4pt}
\begin{tabular}{p{0.26\columnwidth} p{0.23\columnwidth} p{0.23\columnwidth} p{0.16\columnwidth}}
\toprule
\textbf{Validation target} & \textbf{Model} & \textbf{Grouping variable} & \textbf{ROC-AUC} \\
\midrule
Held-out configurations of represented tests & L2 logistic regression & Configuration hash & 0.54 \\
Held-out configurations of represented tests & Random forest & Configuration hash & 0.78 \\
Held-out test identities & L2 logistic regression & Test ID & 0.53 \\
Held-out test identities & Random forest & Test ID & 0.55 \\
\bottomrule
\end{tabular}
\caption{Surrogate-model performance under configuration-held-out and task-held-out validation. Test identity was included as a feature only in the configuration-held-out analysis.}
\label{tab:surrogate_generalization}
\end{table}

\begin{figure*}
\centering
\includegraphics[width=\linewidth]{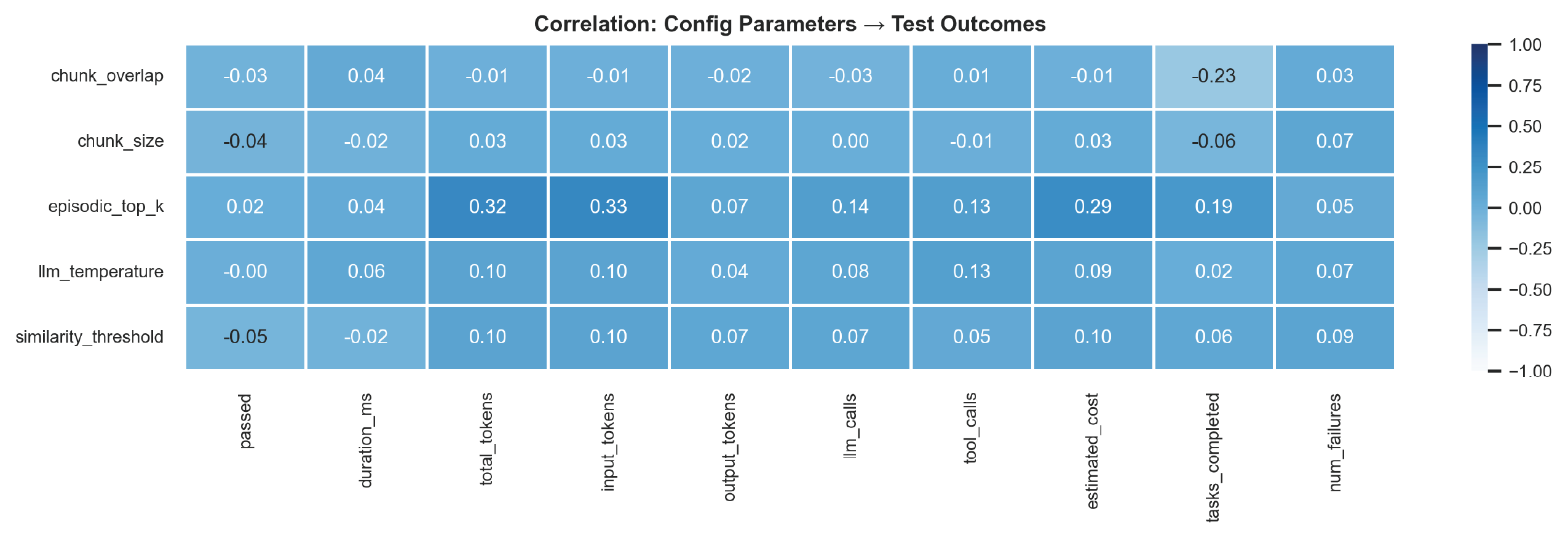}
\caption{Pearson correlations between selected configuration variables and benchmark outcomes.}
\label{fig:hyperparam_efficiency}
\end{figure*}

Figure~\ref{fig:hyperparam_efficiency} summarizes the configuration-held-out surrogate analysis. The correlations and parameter sweeps show that most individual configuration variables had weaker relationships with pass rate than test identity. Configuration variables more consistently affected duration, token use, cost, and the number of actions taken during a workflow.

\subsection{RAG retrieval auditing identified problematic context but did not define a general pruning rule}\label{sec:knowledge_pruning}
\begin{figure*}
\centering
\includegraphics[width=1\linewidth]{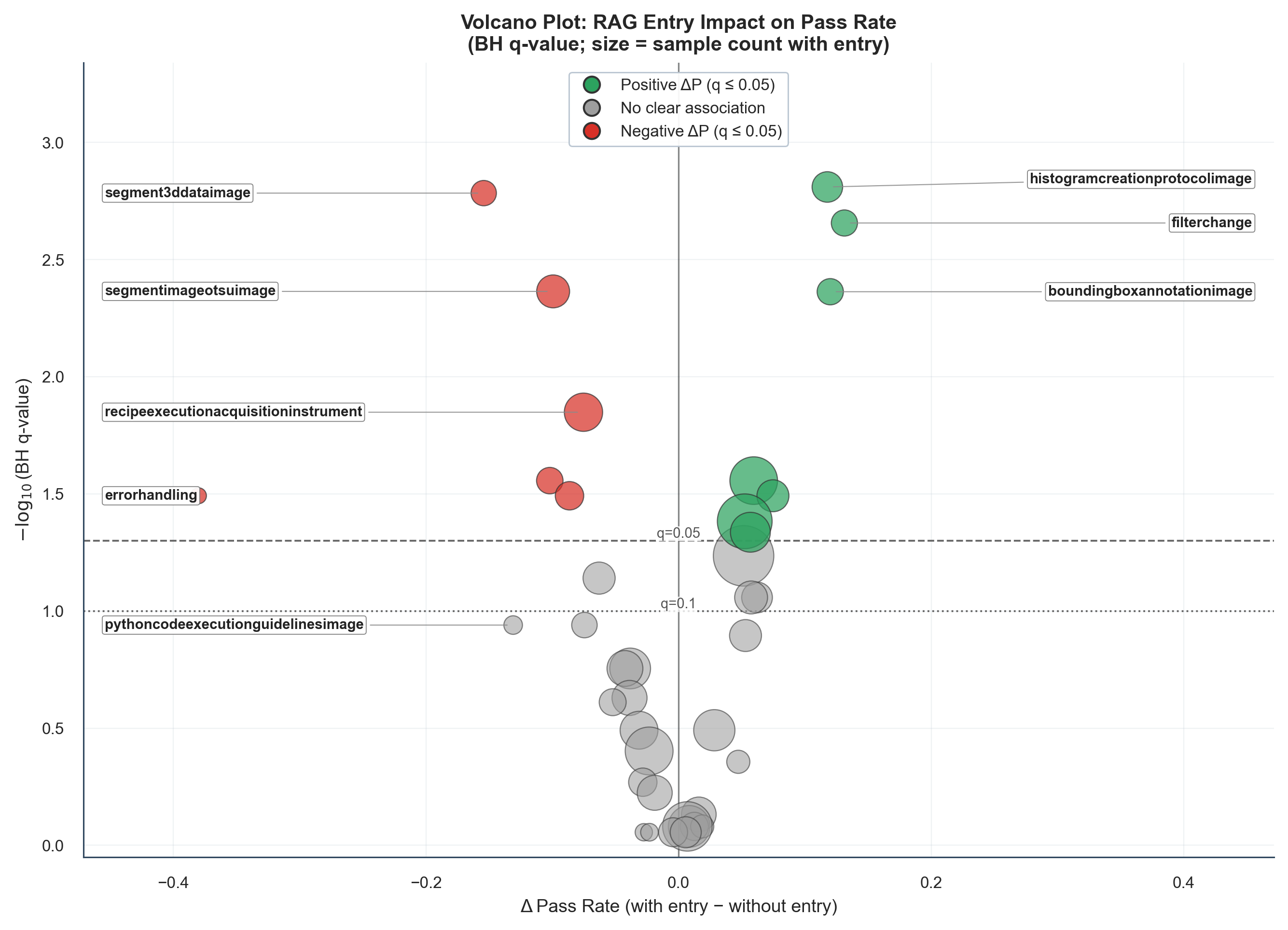}
\caption{Association between retrieval of individual RAG entries and composite benchmark pass rate. The horizontal axis shows the change in pass rate associated with retrieval of an entry, and the vertical axis shows statistical significance after Benjamini--Hochberg adjustment. Point size corresponds to the number of recorded retrievals for the entry. The associations are diagnostic and do not represent causal effects of retrieval.}
\label{fig:RAG_volcano_plot}
\end{figure*}

Every retrieved knowledge fragment was tagged with a stable identifier and linked to the run in which it appeared. This made it possible to compare outcomes for runs in which an entry was retrieved with outcomes for runs in which it was not retrieved. Fisher exact tests followed by Benjamini--Hochberg correction identified entries with adjusted $q$-values associated with higher or lower pass rates. The resulting volcano plot is shown in Figure~\ref{fig:RAG_volcano_plot}.

Several negatively associated entries were also clearly undesirable when inspected manually. For example, \texttt{routingdecision} and \texttt{errorhandling} described supervisor delegation behavior that was irrelevant to a one-agent graph and unnecessary in a two-agent graph. These entries were removed because they were stale or incompatible with the active software.

The more surprising result is the large majority of embedded documents that did not strongly correlate with benchmark task success. The majority of embedded documents fall below the $q = 0.10$ line and do not have a large absolute $\Delta \text{Pass}$ rate. Some of these entries included workflow documents that directly relate to defined benchmark tasks. Many agent configurations were able to successfully pass these benchmarks with similar pass rates without retrieving these documents. Retrieving these documents increased context length and workflow duration, without significantly improving agent behavior. 

The statistical associations were nevertheless observational. Retrieval depended on the task, graph topology, configuration, agent-generated query, and workflow state. Pruning entries with negative marginal associations did not consistently improve performance across configurations. Retrieval identifiers were therefore useful for locating obsolete or contradictory context, but the adjusted associations did not provide an automatic rule for globally improving the RAG database.

\subsection{A one-agent pruned-RAG configuration was the best observed prospective candidate}
The retrospective surrogate analysis was used to select candidate one- and two-agent configurations based on \texttt{Claude Sonnet 4}, temperature $=0.0$, retrieval depth $=1$, and similarity threshold $=0.8$. Each topology was evaluated with the full and pruned RAG databases. A one-agent \texttt{Claude Sonnet 4} configuration with temperature $=0.0$ and no RAG was included as a simple baseline.

Several prospective configurations performed similarly. The highest observed aggregate performance was obtained by the one-agent configuration using the heavily pruned RAG database, and this configuration was also Pareto-optimal when performance and resource use were considered together. The no-RAG one-agent baseline performed approximately as well on the standardized 53-test benchmark while requiring less retrieved context. Multi-agent candidates did not provide a consistent advantage and generally required more tokens, calls, and time.

The prospective result identifies a strong configuration for the workflows tested here, but it does not validate a task-independent global optimizer. A configuration can perform best in a direct comparison while the surrogate used to select it still fails to rank configurations reliably on previously unseen task identities.

\begin{figure*}
\centering
\includegraphics[width=0.85\linewidth]{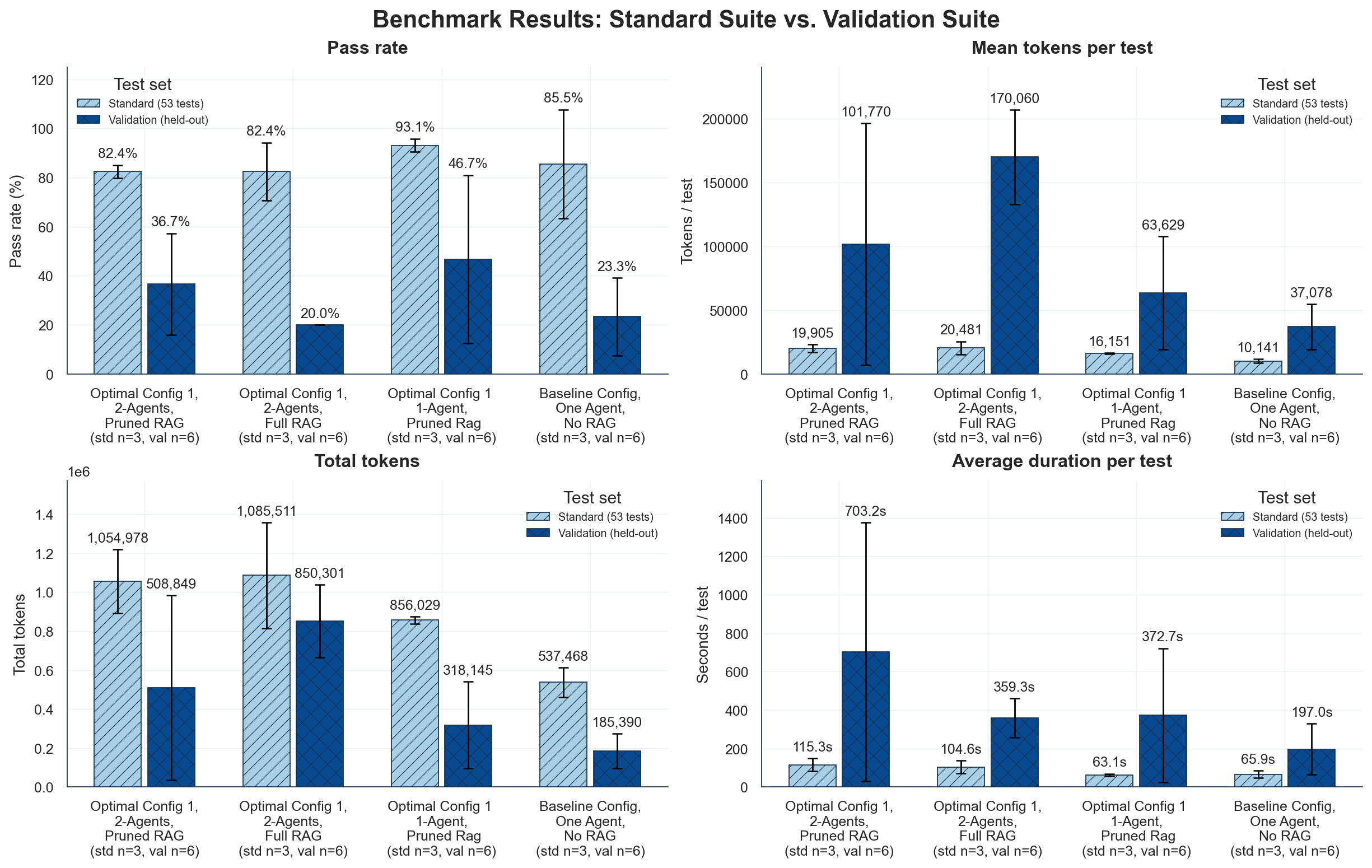}
\caption{Prospective comparison across the standardized 53-test benchmark and five quantitative validation tasks. The comparison includes one- and two-agent \texttt{Claude Sonnet 4} candidates using full or pruned RAG databases and a one-agent no-RAG baseline. The one-agent pruned-RAG configuration produced the highest observed aggregate performance and was Pareto-optimal, while the no-RAG baseline performed approximately as well on the standardized benchmark. Error bars shows the Wilson 95\% confidence interval.}
\label{fig:baseline_comparison}
\end{figure*}

\subsection{The agent completed physical workflows and validation tasks}
The agent was evaluated on five holdout tasks that were not included in the initial 53-test surrogate dataset. Results from both the standard benchmarks and held-out validation benchmarks are seen in Figure \ref{fig:baseline_comparison}. The first task combined instrument setup, sample centering, and source--detector geometry and required the resulting positions to be written to \texttt{setup\_position.json}. The remaining tasks measured the three sample dimensions, optimized exposure for a target sample intensity, created and executed a recipe using the selected settings, and counted defects larger than 1~mm in a reconstructed TIFF stack. The validation criteria are summarized in Table~\ref{tab:validation_benchmarks}, and the complete definitions are reproduced in the system code repository.

\begin{table*}
\centering
\scriptsize
\renewcommand{\arraystretch}{1.18}
\begin{tabularx}{\textwidth}{>{\raggedright\arraybackslash}p{0.13\textwidth} X X >{\raggedright\arraybackslash}p{0.22\textwidth} >{\centering\arraybackslash}p{0.08\textwidth} }
\toprule
\textbf{Validation task} & \textbf{Required output} & \textbf{Reference value or required action} & \textbf{Acceptance criterion} \\
\midrule
Setup, centering, and imaging geometry &
\texttt{setup\_position.json} &
Stage and imaging positions &
Stage X $650\pm100~\mu$m; stage Y $11100\pm500~\mu$m; stage Z $0\pm100~\mu$m; source Z $-87\pm10$~mm; detector Z $50\pm10$~mm \\
Sample dimensions &
\texttt{sample\_size.json} &
$L_x=L_y=L_z=29$~mm &
$L_x=29\pm1$~mm; $L_y=29\pm2$~mm; $L_z=29\pm1$~mm \\
Exposure optimization &
\texttt{optimal\_exposure.json} &
Exposure selected for an average sample-region intensity of approximately 5000 counts &
Optimal exposure $5\pm0.25$~s \\
Recipe creation and execution &
\texttt{TDP\_006\_optimal.rcp} &
Use current stage/source/detector positions and the saved optimal exposure; call recipe modification and execution tools &
Required recipe file exists and acquisition is initiated with the required tool calls \\
Defect count &
\texttt{number\_of\_defects.json} and plots of defect slices &
Five pore regions with diameter larger than 1~mm &
Saved integer count equals 5  \\
\bottomrule
\end{tabularx}
\caption{Holdout physical validation tasks defined in \texttt{validation\_holdout.yaml}. The table reports the aggregate outcome used in the present comparison.}
\label{tab:validation_benchmarks}
\end{table*}

In successful runs, the agent centered an approximately $29\times29\times29$~mm additively manufactured PLA artifact, estimated its dimensions, selected the imaging geometry, created and executed a tomography recipe, reconstructed the volume, and identified defects. Defect segmentation was the least reliable step across the three repeats because it depended on bespoke Python code written during the workflow. Instrument motion and imaging geometry were more reliable because these calculations were performed by deterministic, calibration-aware MCP tools.

\begin{figure*}
\centering
\includegraphics[width=0.75\linewidth]{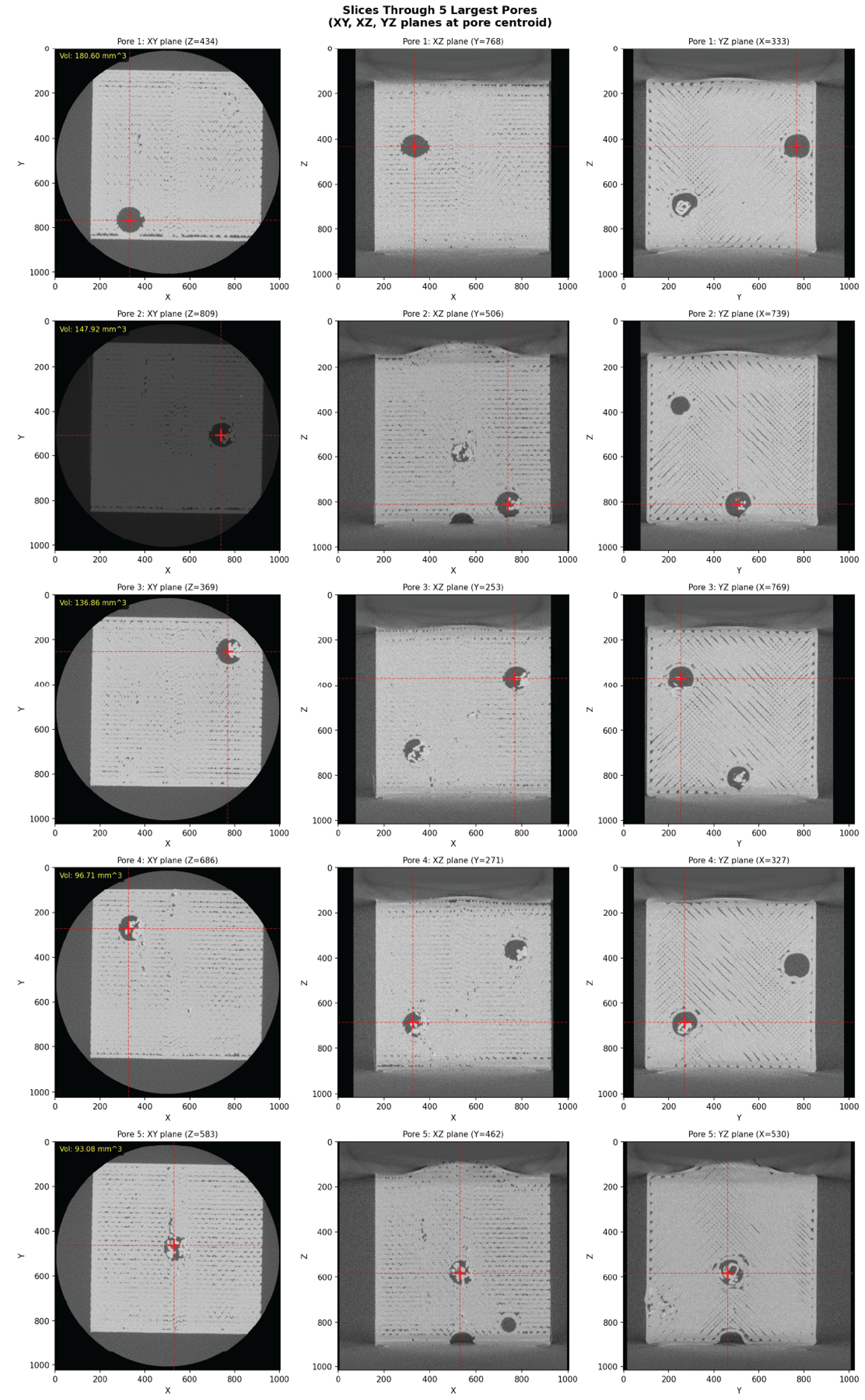}
\caption{Output from a successful physical validation workflow. The agent completed the workflow through reconstruction and defect identification and generated the plotted orthogonal slices during benchmark execution.}
\label{fig:tdp_workflow_example}
\end{figure*}

Figure~\ref{fig:tdp_workflow_example} shows output from a successful execution of the physical validation workflow, illustrating that the system could proceed from instrument setup through quantitative defect identification.

\begin{figure*}
\centering
\includegraphics[width=0.75\linewidth]{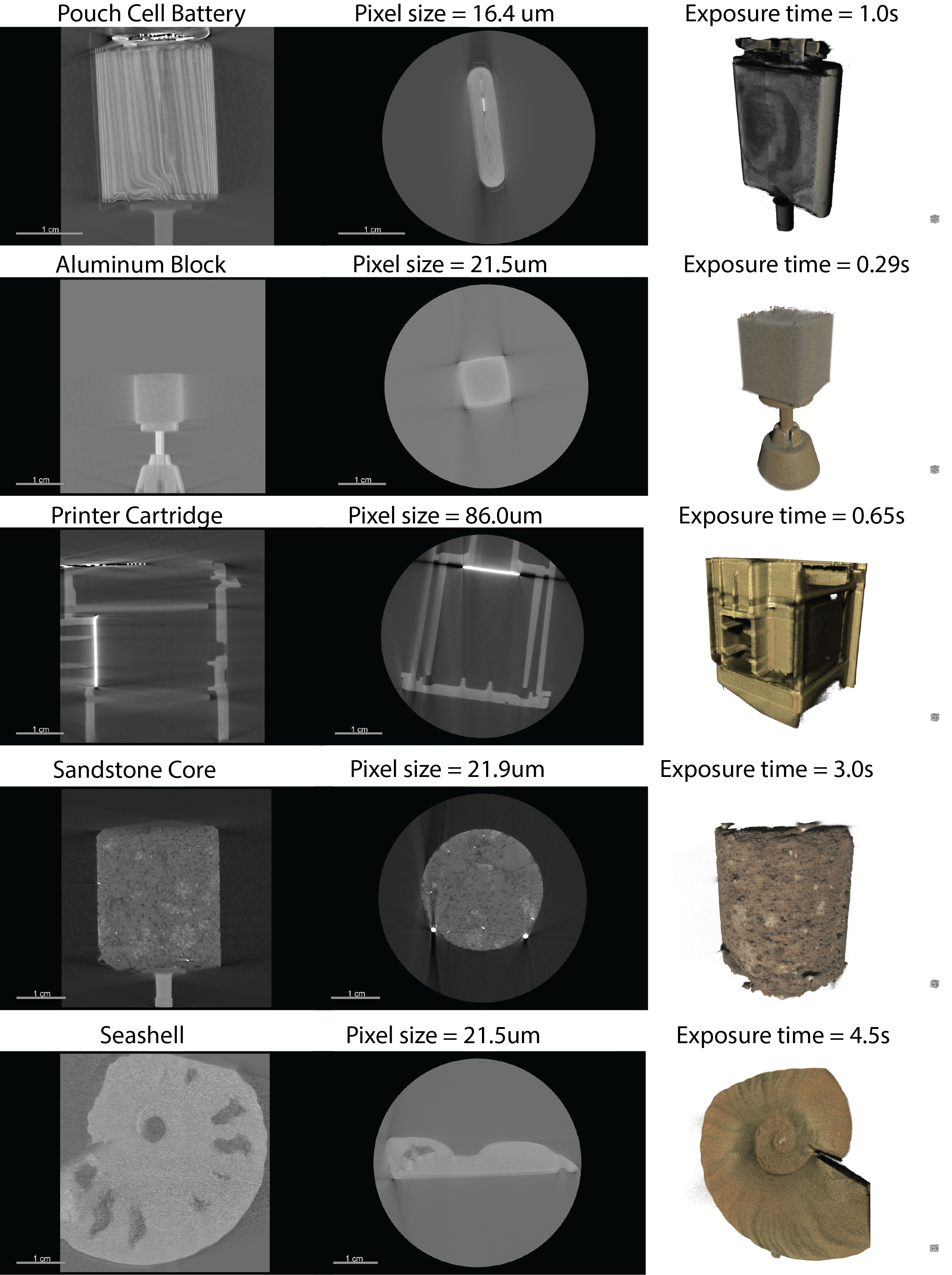}
\caption{Selection of three-dimensional tomography datasets acquired using the same integrated agent prompt. \textbf{Input prompt:} Set the instrument up for general imaging with the 0.4X detector. Center the sample in the field of view. Optimize the exposure time to achieve approximately 10,000 pixel counts in the sample region. Choose a filter based on the average transmission through the sample. Calculate a magnification so that the field of view is 1.25$\times$ wider than the sample. Record the settings in a recipe file named \texttt{SAMPLE\_NAME.rcp} and begin the recipe acquisition.}
\label{fig:demo_sample_suite}
\end{figure*}

The selected system was also used for routine tomography workflows on several samples. A suite of samples surveyed using the optimal configuration is shown in Figure \ref{fig:demo_sample_suite}.  A common prompt required instrument setup with the 0.4X detector, sample centering, exposure optimization, filter selection from measured transmission, magnification selection, recipe creation, and acquisition. The agent completed the major workflow steps for all demonstrated samples and produced data suitable for routine inspection. Conventional tomography artifacts remained in difficult samples, including cone-beam artifacts and streaking caused by highly absorbing copper components. These results show that the absence of a transferable global configuration model is not equivalent to a failure of the microscope agent. The system completed useful physical work; the limitation was the ability of one statistical response surface to rank configurations across unrelated task types.

\section{Discussion}

\subsection{What the benchmark supports}
The main result of this study is that the benchmark was useful for qualification, regression testing, diagnosis, and direct comparison, but it did not support a task-independent model of agent configuration. The configuration-held-out random forest achieved an ROC-AUC of approximately 0.78, which initially suggested that the historical data could be used to optimize the agent. However, test identity was the strongest feature in that model. Once complete tests were held out and test identity was removed, the L2 logistic-regression and random-forest models performed near chance.

This result does not mean that model choice, graph topology, temperature, or RAG parameters have no effect. The direct comparisons clearly showed differences in token use, duration, cost, and failure mode. The prospective comparison also identified a one-agent pruned-RAG configuration that produced the highest observed aggregate performance and was Pareto-optimal. The limitation is narrower: the effects were not stable enough across unrelated tasks to support one global ranking that transferred to a new test identity.

The behavior of the individual tests helps explain this result. Many atomic instrument tests passed under almost every functioning configuration. These tests remain valuable because they verify that the source can be controlled, an axis can move, an image can be acquired, or a known software regression has not returned. They are poor optimization tests because there is little performance variation to model. Other tests were difficult but stochastic, producing both passes and failures without a reproducible relationship to the recorded configuration variables. A benchmark can therefore be useful without being equally informative for every purpose.

The most promising extension is to represent the task explicitly. The task-held-out models knew the LLM, temperature, topology, and RAG settings, but they did not know whether the unseen task required one tool or ten, image analysis, code generation, file creation, a quantitative measurement, or a safety-critical action. Future models could include descriptors such as workflow family, required tools, expected sequence length, multimodal input, quantitative versus trace-based endpoint, and safety constraints. Such a model would ask which configuration is appropriate for a class of tasks rather than assuming that one configuration is best for all tasks.

The direct architecture comparison still provides a useful engineering result. Multi-agent systems did not show a consistent pass-rate advantage and generally required more tokens, calls, and time. Every additional agent adds another context window, another model call, and another coordination boundary. These costs may be justified for long workflows that benefit from explicit planning or specialist tools, but additional agents should not be assumed to improve performance simply because the graph is more elaborate. In this study, the one-agent configurations were consistently difficult to beat.

The prospective comparison reinforces this point. The one-agent pruned-RAG configuration had the best observed aggregate performance, but several candidates were close and the no-RAG one-agent baseline performed approximately as well on the standardized tests. This does not make the prospective experiment uninformative. It identifies a strong practical configuration and shows that a simple baseline should remain part of every comparison. It also prevents a small difference between tested candidates from being interpreted as evidence for a universal optimization rule.

\subsection{Retrieval auditing and tool design}
RAG retrieval identifiers were useful because they made the context window auditable. The logged entries exposed obsolete instructions, references to tools that no longer existed, and documents written for a graph topology that was not active. Removing these entries was justified as software maintenance. However, retrieval was not randomly assigned, and entries associated with failure were often retrieved because of the task, query, or workflow state. The adjusted $q$-values therefore identified entries for inspection, not entries that could be deleted automatically.

There is also confounding context provided both by RAG databases and by the MCP tools. The Model Context Protocol enables the agent to see both the arguments of tools as well as a short text description of the tool. Often, this short text description includes information about when the tool should be used. Thus, having both a RAG entry which describes what tool to call during a specific workflow \textbf{and} having a description in the MCP tool of when to use the tool is redundant. The context provided by tools through MCP can be used to reduce the amount of workflow-related context required in the RAG database.

The failure of global pruning does not reduce the value of retrieval logging. Poor context can degrade agent performance, particularly when irrelevant or incorrect information is injected into the context window \cite{XuRetreivalContextWindow,Cejas2025HallucinationRiskRetrieval}. Stable identifiers make it possible to reproduce the context seen during a failed run and compare that context against successful runs. Historical workflows can be audited in the same way. Prior successful workflows may help an agent complete a similar task \cite{Cherukara2026,JohnsonAugustPrototype,wall2025temagentenhancingtransmission}, but they can also cause the system to copy a previous procedure too closely or retrieve a workflow that is only superficially similar.

The quantitative validation tests also showed that some improvements should be made at the tool level rather than through another global hyperparameter sweep. The least reliable step was free-form defect segmentation written in Python during the workflow. In contrast, stage corrections and magnification calculations were performed by deterministic, calibration-aware MCP tools and were more reproducible. When a scientific operation can be bounded, validated, and reused, it is generally better to expose it as a deterministic tool than to ask the LLM to reconstruct the calculation each time. Free-form code remains useful for new analyses and figure generation, but repeated scientific operations should be moved into tested software as the system matures.

MCP also improved traceability. Earlier instrument agents often relied on LLM-written code to access a microscope API\@. Code generation is flexible, but it is difficult to constrain and inspect. A curated tool surface records the sequence of instrument actions and allows limits, safety checks, and refusal behavior to be enforced outside of the LLM\@. The agent can still decide which tool to use and in what order, but it cannot bypass the instrument interface. This separation is especially important for autonomous scientific systems that interact with physical hardware \cite{Tang2025NatureCommSafety}.

\subsection{Implications for microscopy benchmarks}
Agentic microscopy is still an early field, and published systems differ substantially in graph topology, tool access, memory, and the division of responsibility among agents. These architectural differences make direct trace-to-trace comparison difficult. A common benchmark can still compare whether the systems complete the same scientific task, even when they use different internal workflows.

No single endpoint is sufficient for every benchmark purpose. Text matching is appropriate when the desired output is text, and routing accuracy is appropriate when the task is to delegate work. Mathur et al., for example, used word-error rate to evaluate a classifier agent that routed synchrotron tasks \cite{Mathur2025VISION}. For a microscope controller, the final quantitative measurement is the highest-order scientific endpoint, but qualification and regression tests remain necessary. A system that measures the correct feature but exceeds a safety limit is not acceptable, and a system that passes a smoke test has not necessarily demonstrated quantitative accuracy.

The composite pass rate used here combined all correctness requirements defined for a test, including quantitative output, required artifacts, expected actions, and operational limits. This provided a strict headline outcome. The underlying checks were still needed to determine why a run failed. Scientific correctness, trace compliance, and operational efficiency should therefore remain separately visible in the benchmark record even when they are combined into one pass/fail result.

Panigrahi et al.\ distinguish between benchmarking LLM knowledge and benchmarking the performance of the complete agent system \cite{panigrahi2026,ai2026}. This distinction is particularly important for instrument-driving agents. A plausible final response does not establish that the microscope moved correctly, and an unexpected trace does not necessarily mean that the final measurement was wrong. Standard samples with known dimensions, defects, positions, or image features could provide a common quantitative basis for evaluation, similar to the use of imaging phantoms for microscope characterization \cite{Postek1994SEM,Groenewald2016CTPhantom,Liam2023ImagePhantoms}.

Several limitations affect the present analysis. The configuration dataset was historical and unbalanced rather than a complete factorial experiment. Model, prompt, framework, code, software, and hardware versions changed during the study. Runs were collected using a simulator and multiple ZEISS Versa microscope models. Repeated runs of the same task and configuration were not fully independent. Claude Opus 4.5 was also affected by an implementation incompatibility in the multi-agent graphs. These factors may contribute to apparent configuration effects and limit the precision of feature rankings.

The benchmark nevertheless exposed difficult workflows, architecture overhead, operational failure modes, stale RAG context, and weaknesses in free-form analysis. It also supported regression testing and physical validation. What it did not provide was evidence that the same global configuration ranking would transfer to any unseen microscopy task. Recognizing this boundary narrows the next engineering problem: preserve the benchmark for qualification and diagnosis, and develop task-aware models or targeted experiments when a particular workflow needs to be improved.

\section{Conclusion}
We developed a benchmark-linked framework for evaluating agent configuration, operational behavior, and RAG retrievals in an agentic X-ray microscope controller. The dataset contained 53 benchmark tests, 105 configurations, 1,949 test runs, and 49,109 retrieval events. Direct comparisons showed that graph topology and LLM choice changed token use, duration, cost, and failure mode. In the prospective comparison, a one-agent configuration using a heavily pruned RAG database produced the highest observed aggregate performance and was Pareto-optimal. A one-agent no-RAG baseline nevertheless performed approximately as well on the standardized benchmark tests, showing that additional context and architectural complexity did not provide a consistent advantage.

The surrogate-modeling results place an important limit on what can be concluded from the benchmark. A random forest achieved an ROC-AUC of approximately 0.78 when configurations were held out but known test identities remained represented in training. When complete tests were held out, the configuration signal did not transfer. The benchmark is therefore useful for qualification, regression testing, diagnosis, and direct comparison of defined workflows, but the present heterogeneous test suite does not support a task-independent global configuration model.

The agent still completed instrument setup, sample centering, exposure and filter selection, geometric calculations, recipe generation, tomography acquisition, and quantitative measurements on physical samples. Future optimization should focus on task-aware configuration selection and on replacing repeated free-form operations with validated deterministic tools. This approach preserves the practical value of benchmarking without assuming that one agent configuration will be optimal for every scientific task.

\section*{Code and Data Availability}
The benchmark records, preprocessing scripts, surrogate-modeling code, and plotting notebooks are available at \texttt{github.com/natertott/agentic\_microscopy\_benchmarks\_XRM}. A PDF of the Supplemental Information can also be found in that repository.

\section*{Acknowledgements}
The authors thank Roland Salzer and Simon Franchini of Carl Zeiss Corporate Research and Technology for helpful discussions. The authors also thank the ZEISS X-ray Microscopy engineering, advanced development, field-of-business, and marketing teams for their support of this work.

\section*{Author Contributions}
\textbf{Ian Abshire}: Software, investigation, data curation, and writing--review and editing.\\
\textbf{Nathan S. Johnson}: Conceptualization, methodology, software, investigation, formal analysis, data curation, visualization, and writing--original draft, review, and editing.

\section*{Funding}
This work was funded by Carl Zeiss Research Microscopy Solutions.

\section*{Declaration of Competing Interests}
The authors are employees of Carl Zeiss Research Microscopy Solutions. The work used ZEISS instrumentation and software. The authors declare no additional competing interests.

\section{Methods}
\subsection{System architecture and instrumentation}
The agentic microscope controller was implemented using LangChain and LangGraph. All agents followed a Reasoning and Acting (ReAct) workflow \cite{yao2023react}. In the one-agent graph, a single worker planned the workflow and called all available tools. The two-agent graph used a supervisor to develop and delegate a plan to one worker. The three-agent graph used a supervisor and separate instrument and image-analysis workers. The graph topologies are shown in Figure~\ref{fig:system_setup}.

Microscope control was provided through the Model Context Protocol (MCP) \cite{modelcontextprotocol2025,anthropic2024mcp}. The LangGraph controller acted as an MCP client and sent structured tool calls to six Python MCP servers. The servers controlled the microscope, created and executed acquisition recipes, performed geometry and optics calculations, processed images, read and wrote files, and executed custom Python code. The LLM did not directly call the vendor API or manipulate the microscope workstation outside of these tools.

The system was tested using simulated microscope-control software and ZEISS Versa 515, 630, and 730 X-ray microscopes at different stages of development. The simulator exposed the same Python 3.11 instrument API used by the physical system and was used for rapid software testing and large benchmark sweeps. Final physical validation and demonstration workflows were performed on a ZEISS Versa 730 using version 3.0 of the ZEN NavX interface and its Python 3.11 API.

\subsection{Study design}
The study was organized into six stages. First, one-, two-, and three-agent topologies were screened on a common subset of 15 benchmark tests using the five LLMs included in the study. Second, the broader 53-test benchmark was used to evaluate a larger set of agent configurations and RAG parameters. Third, L2-regularized logistic-regression and random-forest surrogate models were trained under both configuration-held-out and task-held-out validation schemes. Fourth, RAG retrieval identifiers were linked to benchmark outcomes, and entries associated with poor performance were reviewed and selectively removed. Fifth, surrogate-selected one- and two-agent configurations using full and pruned RAG databases were evaluated prospectively against a simple one-agent no-RAG baseline. Finally, selected configurations were used for quantitative validation tasks and complete physical microscopy workflows.

The same benchmark runner was used throughout the study. It recorded the configuration snapshot, test outcome, tool and agent traces, duration, token use, estimated cost, LLM calls, tool calls, created files, errors, and all RAG retrieval events. The design was historical and unbalanced rather than a complete randomized factorial experiment; not every configuration was run on every test and the number of repeats varied across the dataset.

\subsection{Context construction and agent state}
Each agent received a role-specific system prompt, information describing the workspace and available files, any retrieved RAG context, and the current shared \texttt{AgentState}. The state contained the bounded message history, task list, current working directory, created files, structured tool results, retry count, shared data, and compressed summaries of older messages. The full state schema is provided in Supplementary Table 1.

Figure~\ref{fig:context_window} summarizes the context and execution flow. In the one-agent topology, planning and tool execution occurred within one ReAct loop. In the two- and three-agent topologies, the supervisor maintained the workflow plan and delegated tasks to one or more workers with different prompts, tool permissions, and RAG collections.

\begin{figure*}[t]
\centering
\includegraphics[width=0.9\linewidth]{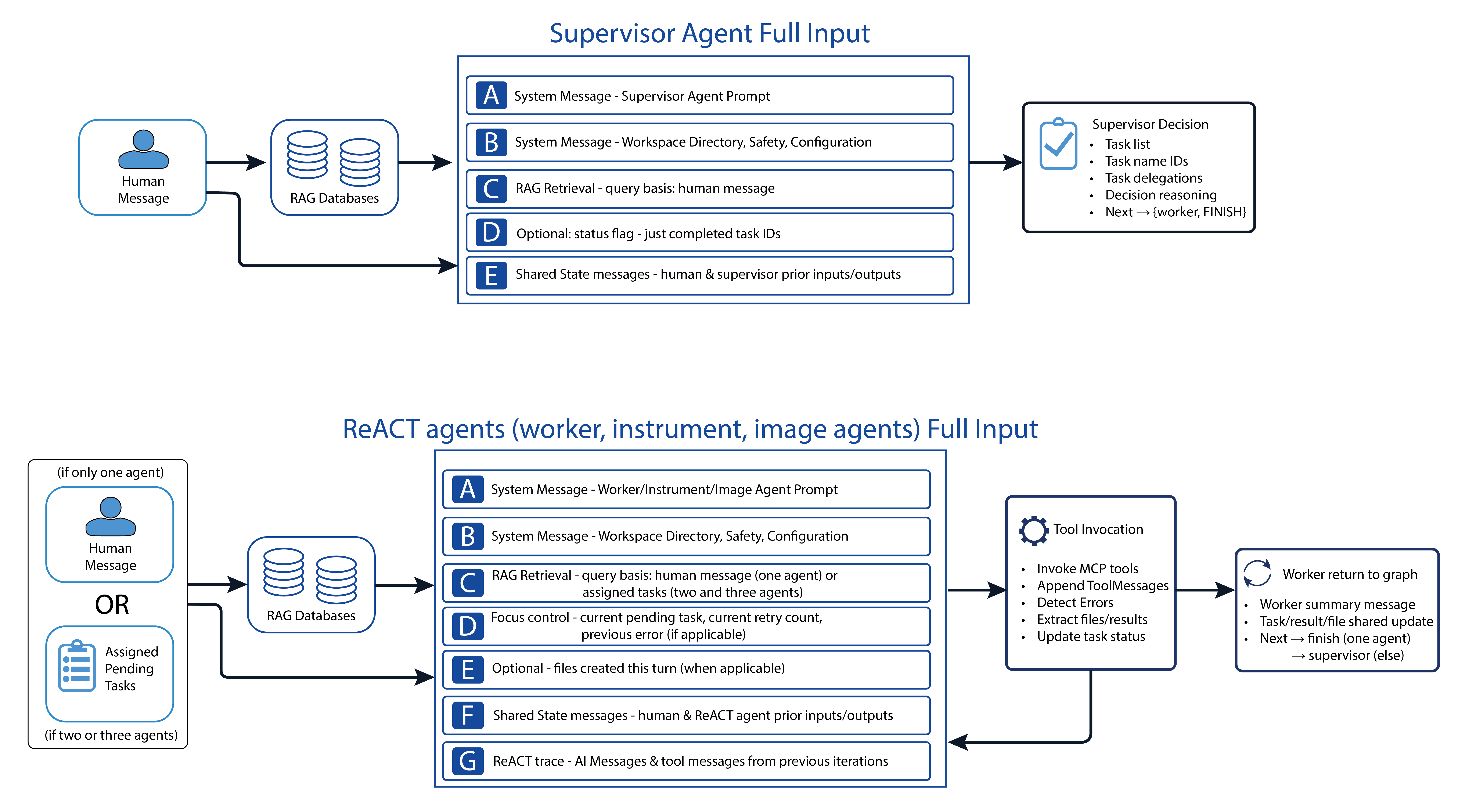}
\caption{Context construction and execution flow for supervisor and ReAct worker agents. The one-agent topology combines planning and execution in one context window. The two-agent topology uses a supervisor and one worker, while the three-agent topology uses a supervisor and separate instrument and image-analysis workers.}
\label{fig:context_window}
\end{figure*}

\subsection{Configuration parameters and LLMs}
The configuration variables represented in the benchmark dataset are summarized in Table~\ref{tab:hyperparameter_list}. A value of zero for either retrieval-depth parameter disabled retrieval from that database. The RAG database was also evaluated in its full, pruned, and disabled states. Because the study design evolved over time, the table describes the range of values represented in the dataset rather than a complete factorial grid.

Five LLMs were evaluated throughout the study: \texttt{Claude Sonnet 3.7}, \texttt{Claude Sonnet 4}, \texttt{Claude Opus 4.5}, \texttt{GPT-4.1}, and \texttt{GPT-5.1}. Model names are reported as display names for readability; the exact API model identifiers and configuration snapshots are retained in the benchmark records.

\begin{table*}[t]
\centering
\small
\setlength{\tabcolsep}{4pt}
\renewcommand{\arraystretch}{1.15}
\begin{tabularx}{\textwidth}{l l l l X}
\toprule
\textbf{Parameter} & \textbf{Type} & \textbf{Values / range} & \textbf{Default} & \textbf{Description} \\
\midrule
number of agents & categorical & 1, 2, 3 & 1 & Number of LLM agents in the graph topology. \\
LLM model & categorical & five models & \texttt{GPT-4.1} & Model used by the active agent or agents. \\
LLM temperature & float & 0.0--1.0 & 1.0 & Sampling temperature used for model generation. \\
episodic \texttt{top\_k} & integer & 0--5 & 2 & Number of procedural or prior-workflow entries retrieved. \\
contextual \texttt{top\_k} & integer & 0--10 & 3 & Number of sample- or task-context entries retrieved. \\
chunk size & integer & 500--2000 & 1000 & Number of characters or tokens represented in an embedded text chunk, as defined by the active preprocessing pipeline. \\
chunk overlap & integer & 50--500 & 200 & Overlap between adjacent embedded chunks. \\
similarity threshold & float & 0.1--1.5 & 0.95 & Minimum retrieval-similarity criterion. \\
RAG database state & categorical & none, full, pruned & full & Knowledge database made available to the agent. \\
maximum context tokens & integer & 5000--50000 & 20000 & Maximum context budget applied by the message-management policy. \\
recursion limit & integer & 20--200 & 100 & Maximum graph recursion depth before forced termination. \\
\bottomrule
\end{tabularx}
\caption{Agent configuration variables represented in the benchmark dataset. The sampled design was unbalanced, and not every combination of values was evaluated.}
\label{tab:hyperparameter_list}
\end{table*}

\subsection{Benchmark definitions and composite pass rate}
The benchmark contained 53 tests grouped into seven suites. The suite structure, outcome types, tracked metrics, and six-test end-to-end golden suite are summarized in Supplementary Tables 3-5. The full test definitions include the prompt, required outputs, tool or agent requirements, file checks, quantitative tolerances, and operational limits for each test.

The headline pass rate used a composite definition. A run passed only when all checks defined for that test were satisfied. Depending on the test, these checks included a correct physical or numerical outcome, required tool or agent calls, expected text or file contents, artifact creation, and limits on duration, LLM calls, tool calls, or timeout. The individual checks were retained in the run record so that failures caused by scientific error, trace deviation, or operational limits could be analyzed separately even though the headline outcome was binary.

\subsection{Dataset construction}
The primary data source was the suite-level JSON output generated by the benchmark framework. Each suite file contained a suite identifier, suite name, configuration snapshot, start and end times, duration, counts of passed, failed, skipped, and errored tests, and the aggregate suite pass rate. The configuration snapshot recorded the model, graph topology, RAG parameters, message-management settings, available tools, test settings, cost settings, and file paths.

Each test record contained the test identifier, test name, binary outcome, timing data, configuration hash, failed checks, error messages, input and output token use, LLM-call and tool-call counts, tool names and invocation records, agent visitation sequence, estimated cost, RAG retrieval events, final response, created files, message trace, and workspace path. Estimated cost was calculated from recorded token use and the model-specific pricing configured in the benchmark framework.

RAG retrieval records were nested within each test run. Each event included the receiving agent, query type and text, number of returned entries, context-token count, retrieval latency, entry metadata, stable entry identifiers, timestamp, and workflow step. A single run could contain multiple retrieval events, and each event could return several entries.

\subsection{Surrogate design matrices}
Two design matrices were generated. The pre-run matrix contained only information known before execution, including test identity and agent configuration. It was used to estimate the probability that a proposed configuration--test pair would pass. The RAG-aware matrix added multi-hot indicators for entries retrieved during execution. This second matrix was used for diagnosis rather than prospective prediction because retrieval behavior was not known before the run began.

The binary target was the composite pass outcome. The surrogate models were intended to approximate the observed response surface and identify reproducible associations. They were not treated as replacements for direct benchmark testing.

\subsection{L2-regularized logistic regression}
An L2-regularized logistic-regression classifier was used as an interpretable linear baseline. For feature vector $x$, the predicted probability of passing was

\begin{equation}
P(\mathrm{pass}\mid x)=\frac{1}{1+\exp[-(\beta_0+\boldsymbol{\beta}^{T}x)]}.
\label{eq:logistic_model}
\end{equation}

The L2 penalty was used to stabilize the model in the presence of correlated configuration variables and a design matrix with many encoded features. Coefficient signs were interpreted as conditional associations with the log odds of passing, not as causal effects.

\subsection{Random-forest classification}
A random-forest classifier was used as a nonlinear surrogate. The model combined predictions from an ensemble of decision trees and could represent nonlinear responses and interactions between configuration variables. Feature importance was used as a screening measure for variables that contributed to the fitted trees. Because impurity-based importance can favor variables with many levels and can distribute importance across correlated predictors, the resulting rankings were interpreted together with logistic-regression results, descriptive summaries, and the cross-validation design.


\subsection{Cross-validation and generalization tests}
Two grouped validation schemes were used. In the configuration-held-out analysis, all runs with the same configuration hash were assigned to the same fold. Test identity remained in the design matrix, and the same benchmark test could therefore appear in both training and validation folds under different configurations. This analysis measured interpolation to new configurations of already represented tasks.

In the task-held-out analysis, all runs belonging to the same test identity were assigned to the same fold, and test-identity features were removed from the design matrix. Complete tasks were therefore absent from training when their runs were evaluated. This analysis measured whether configuration parameters alone supported generalization to previously unseen benchmark tasks. Grouped cross-validation used up to five folds, limited by the number of available groups and observations in the minority class. ROC-AUC was used as the primary discrimination metric.

A sensitivity analysis repeated the task-held-out evaluation after removing tests that passed for nearly every configuration. This tested whether the large number of qualification and smoke tests was masking a configuration signal among the more difficult tasks.

\subsection{RAG retrieval association and pruning}
RAG-entry associations were evaluated at the run level. For each entry, a $2\times2$ contingency table compared pass and fail outcomes for runs in which the entry was retrieved with runs in which it was not retrieved. Fisher's exact test was used to calculate an odds ratio and $p$-value. Benjamini--Hochberg correction was applied across entries, and adjusted $q$-values were used for multiple-comparison control. Entries were classified by the sign of the pass-rate difference and a threshold of $q\leq0.05$.

These tests were observational. Retrieval was not randomly assigned and depended on the task, configuration, agent-generated query, graph topology, and workflow state. Entries associated with poor performance were therefore reviewed manually before pruning. Obsolete, contradictory, or topology-inappropriate entries were removed to create a pruned database, which was then evaluated directly against the full database and a no-RAG baseline.

\subsection{Prospective configuration comparison}
Candidate configurations were selected from the retrospective surrogate analysis and evaluated prospectively on the standardized benchmark and the quantitative validation tasks. The comparison included one- and two-agent versions using full and pruned RAG databases. A one-agent \texttt{Claude Sonnet 4} configuration with no RAG was included as a simple baseline. Performance was compared using composite pass rate, token use, duration, cost, LLM calls, and tool calls. Pareto optimality was used to identify configurations for which no other tested configuration improved one objective without worsening another.

\subsection{Physical validation and demonstration workflows}
Five quantitative validation tasks were used that were not included in the initial 53-test surrogate dataset. The tasks evaluated sample centering, exposure selection, sample-dimension measurement, magnification geometry, recipe creation, and defect counting. Ground truth was supplied by known stage positions, expert-selected acquisition conditions, caliper measurements, calculated source and detector positions, and expert defect counts. Each validation task was repeated three times. The criteria are summarized in Table~\ref{tab:validation_benchmarks}.

The selected agent was also used to perform complete tomography workflows on a set of routine samples. A common prompt instructed the system to configure the 0.4X detector, center the sample, optimize exposure and filter selection, calculate a field of view relative to the measured sample dimensions, create a recipe, and begin acquisition. These demonstrations were used to verify that the benchmarked system could complete integrated physical workflows; they were not included in surrogate-model training.

\bibliographystyle{unsrt}
\bibliography{bib}

\end{document}